\documentclass[letterpaper,10pt,journal,twoside]{IEEEtran}
\usepackage{graphics} 
\usepackage{epsfig} 
\usepackage{array}
\usepackage{float} 
\usepackage[T1]{fontenc}
\usepackage{xcolor,soul}
\usepackage{bm}
\usepackage{pifont}

\usepackage[caption=false,font=footnotesize]{subfig}
\usepackage{etoolbox}
\AtBeginEnvironment{tabular}{\footnotesize}
\usepackage{amsfonts,amssymb, amsmath}
\usepackage{cases}
\usepackage{graphicx}
\usepackage{booktabs}
\usepackage{multirow}
\usepackage{multicol}
\usepackage{hyperref}
\usepackage{capt-of}
\hypersetup{
	colorlinks=true,
	linkcolor=black,
	filecolor=black,
	urlcolor=black,
	citecolor=black,
	breaklinks=true,
	pdftitle={Learning Terrain-Aware Whole-Body Control for Perceptive Legged Loco-Manipulation},
	pdfauthor={Sikai Guo, Yudong Zhong, Guoyang Zhao, Botao Dang, Zhihai Bi, and Jun Ma},
}
\usepackage{mathtools}
\newcommand{\rev}[1]{#1}
\newcounter{RNum}
\renewcommand{\theRNum}{\arabic{RNum}}
\newcommand{\Remark}{\noindent\textit{\textbf{Remark}~\refstepcounter{RNum}\textbf{\theRNum}: }}
\soulregister\Remark7

\begin{document}

\title{Learning Terrain-Aware Whole-Body Control for Perceptive Legged Loco-Manipulation}

\author{Sikai~Guo$^{1}$, Yudong~Zhong$^{1}$, Guoyang~Zhao$^{1}$, Botao~Dang$^{1}$, Zhihai~Bi$^{1}$, and~Jun~Ma$^{1}$%
\thanks{Manuscript received: April 30, 2026; Revised: July 30, 2026; Accepted: August 30, 2026.}%
\thanks{This paper was recommended for publication by Editor Olivier Stasse upon evaluation of the Associate Editor and Reviewers' comments.}%
\thanks{$^{1}$Sikai Guo, Yudong Zhong, Guoyang Zhao, Botao Dang, Zhihai Bi, and Jun Ma are with the Robotics and Autonomous Systems Thrust, The Hong Kong University of Science and Technology (Guangzhou), Guangzhou 511453, China (email: {sguo837@connect.hkust-gz.edu.cn, yzhong369@connect.hkust-gz.edu.cn, gzhao492@connect.hkust-gz.edu.cn, bdang695@connect.hkust-gz.edu.cn, zbi217@connect.hkust-gz.edu.cn, jun.ma@ust.hk}).
        {\tt\footnotesize }}%
\thanks{Digital Object Identifier (DOI): see top of this page.}%
}

\markboth{IEEE Robotics and Automation Letters. Preprint Version. Accepted August, 2026}%
{Guo \MakeLowercase{\textit{et al.}}: Learning Terrain-Aware Whole-Body Control for Perceptive Legged Loco-Manipulation}

\IEEEaftertitletext{
\vspace{-3.0em}
\noindent\begin{center}
\includegraphics[width=0.9\linewidth]{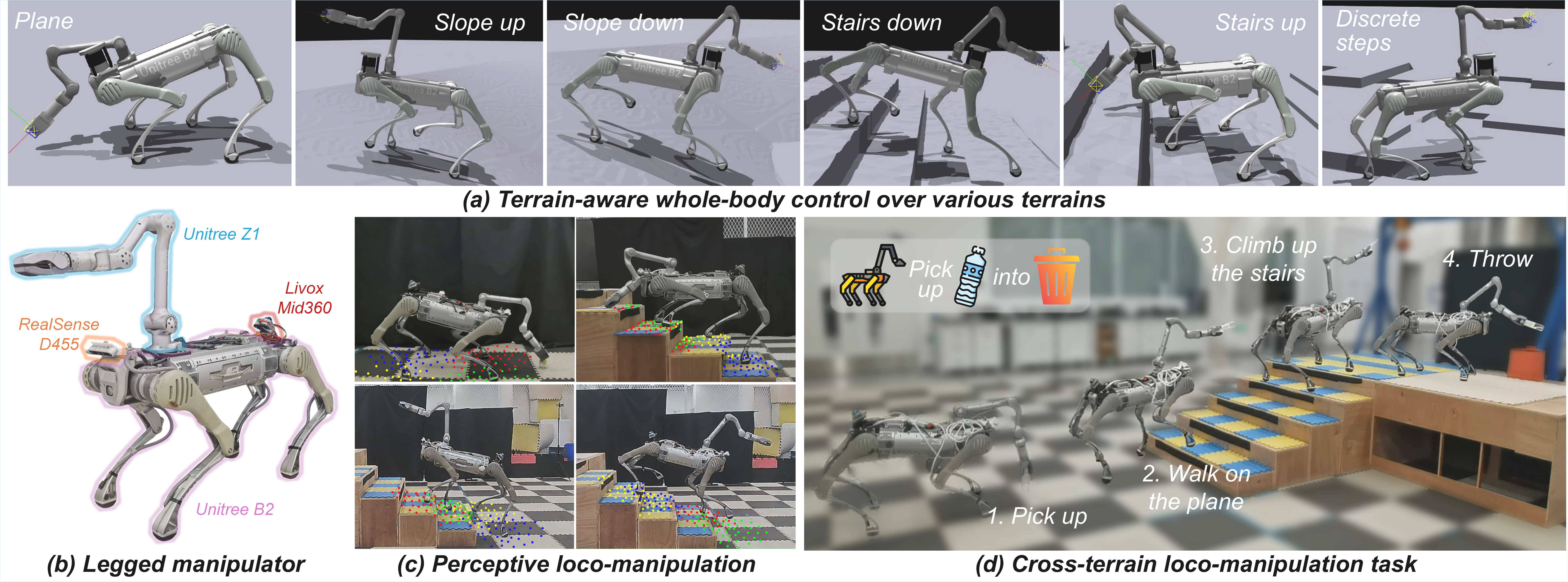}
\vspace{-8pt}
\captionof{figure}{
TA-WBC is a terrain-aware whole-body controller for perceptive legged loco-manipulation over diverse challenging terrains, including slopes, stairs, and steps. With terrain exteroception, it enables legged manipulators to actuate whole-body joints for expanding workspace, while maintaining collision-free cross-terrain traversal capabilities.
}
\label{cover-figure}
\end{center}
\vspace{-5pt}
}

\maketitle

\begin{abstract}

Legged manipulators integrate exceptional terrain adaptability along with mobile manipulation capabilities, which make them highly promising for deployment in human-centric environments.
By coordinating the control of both legs and arms, a whole-body controller can significantly expand the operational workspace of legged manipulators. 
However, many existing whole-body controllers primarily depend on proprioception and do not incorporate the critical exteroception required for effective terrain topology perception. This limitation can hinder their ability to adapt to varying environmental conditions and navigate complex terrains effectively. 
In this paper, we introduce TA-WBC, a terrain-aware whole-body control framework for legged manipulators, which features a novel RL-based unified policy tailored to whole-body loco-manipulation tasks in various terrains. 
Specifically, we employ a \rev{hierarchical exteroceptive encoder} to extract terrain features, providing an essential basis for the robot to proactively adapt posture and footholds. Furthermore, to facilitate stable cross-terrain loco-manipulation, we propose a novel end-effector sampling method based on the foot contact plane, \rev{decoupling the manipulation target from base height, roll, and pitch variations}. Moreover, a dual-policy distillation module is introduced to integrate expansive whole-body motion with terrain adaptability without catastrophic forgetting. 
The simulation and real-world experiments validate the robustness of our proposed controller, which leads to a larger reachable space, less tracking error, and reduced unexpected stumbles. This unified policy highlights the promising capabilities of legged manipulators in performing loco-manipulation tasks across complex terrains.

\end{abstract}


\IEEEpeerreviewmaketitle

\section{Introduction}

\IEEEPARstart{O}{ver} the past decade, mobile manipulation systems have achieved significant advancements, with a remarkable transition from confined industrial environments to human-centric settings \cite{aloha, mm2}. A typical mobile manipulator generally comprises a mobile base and an onboard robotic arm, allowing the system to move its base while simultaneously performing manipulation tasks with the end effector, thereby extending its effective workspace. 
While there have been some rather promising preliminary works on wheeled manipulators in structured spaces~\rev{\cite{xiong2024adaptive,wutidybot++}}, their mobility is \rev{generally} constrained to planar or flat surfaces, limiting their utility in complex scenarios such as disaster response, construction sites, or multi-level residential areas. Legged manipulators, particularly quadruped robots equipped with robotic arms, offer a compelling solution \cite{eth-wbc-mpc,dwbc}. Essentially, they not only have the inherent capability of traversing unstructured terrains, but also can leverage the redundancy of the floating base to extend the reachable workspace. For instance, they can bend their front legs to reach the ground with the end-effector, or elevate the front body to grasp higher objects\cite{roboduet}.

    
    

Despite this potential, achieving coordinated whole-body control (WBC) for legged manipulators remains inherently challenging. The core difficulty lies in the complex whole-body dynamics and the coupling between the high degree-of-freedom (DoF) system and high-accuracy manipulation tasks. Traditional model-based methods for WBC, such as model predictive control (MPC) \cite{eth-wbc-mpc,2025mpc-wbc}, or hierarchical quadratic programming (QP) \cite{qp1,qp2}, rely on precise dynamic modeling and are often computationally intensive. This feature makes them susceptible to unmodeled external disturbances, and unable to stay robust in unstructured environments. 
Conversely, reinforcement learning (RL) has recently emerged as a powerful alternative for developing robust legged-manipulation controllers. By introducing regularized online adaptation \cite{dwbc} and teacher-student training framework \cite{teacher-student}, the RL-based controller bridges the sim-to-real gap, demonstrating superior disturbance rejection and enhanced whole-body coordination \cite{eth-door-open,jkw-ral}, all without requiring complex dynamic modeling. 

However, most existing whole-body controllers for legged manipulation are developed under the assumption of flat and quasi-horizontal terrain. They are typically reliant on proprioception solely, enabling locomotion without external environmental perception, 
and treating terrain irregularities as unmodeled disturbances that are addressed in a reactive manner \cite{vbc}.
Although recent works like PILOT \cite{pilot} achieves perceptive cross-terrain loco-manipulation for humanoid robots, perceptive whole-body control remains particularly challenging for quadrupedal manipulators, since they lack direct human demonstrations for their non-anthropomorphic morphology.
Overall, there are three main challenges when the legged manipulators are executing loco-manipulation tasks in unstructured environments. First, when traversing terrains like stairs or steps, a blind robot has to first make unexpected contact with the obstacles to sense the disturbance, inevitably causing sudden pose perturbation in the floating base and compromising onboard end-effector tracking accuracy. 
Second, the training of existing works relies on end-effector goal sampling strategies restricted to flat terrains, which fails to achieve continuous stability of the end-effector in cross-terrain loco-manipulation tasks due to \rev{base height, roll, and pitch variations}.
Third, although decoupling perceptive locomotion and blind whole-body control into separate policies can be feasible in certain cases \cite{eth-ee-tracking}, this solution requires a complex high-level decision module or manual operation for policy switching. Generally, a unified controller can address this challenge, but it is difficult to train a single policy integrating whole-body coordination ability with multi-terrain adaptability.

To solve the above open challenges, we present the Terrain-Aware Whole-Body Control (TA-WBC), a unified end-to-end low-level controller for perceptive legged loco-manipulation over various terrains, as shown in Fig. \ref{cover-figure}. 
Specifically, it incorporates a \rev{hierarchical exteroceptive encoder} of convolutional networks and multilayer perceptron (MLP), capable of extracting environmental geometric features around feet to avoid undesired foot contact while traversing terrains. 
Furthermore, to facilitate end-effector tracking accuracy in cross-terrain loco-manipulation, we propose a novel end-effector sampling method in the training period. It constructs a distance-invariant sampling sphere relative to the foot contact plane, which \rev{decouples the end-effector goal from base height, roll, and pitch variations}, \rev{so that the command remains stable for continuous end-effector tracking while the robot is traversing uneven terrains}.
Moreover, to mitigate \rev{catastrophic forgetting of whole-body coordination when a quasi-level loco-manipulation policy is further trained for multi-terrain traversal}, we develop a two-stage learning framework with dual-policy distillation, \rev{in which a cross-terrain locomotion expert and a quasi-level whole-body loco-manipulation expert jointly supervise the student policy TA-WBC during training}.
Compared to the decoupled controller and the conventional blind WBC, TA-WBC demonstrates superior performance by traversing complex terrains while mitigating unexpected foot-end collisions, thereby achieving both high terrain adaptability and precise end-effector tracking.
The main contributions are summarized as follows:
\begin{itemize}
    \item We propose TA-WBC, a unified terrain-aware whole-body control framework \rev{that integrates terrain perception and whole-body coordination into a single policy, enabling legged manipulators to simultaneously traverse diverse terrains and track end-effector commands.}
    \item We propose a novel end-effector sampling approach tailored to \rev{continuous} cross-terrain loco-manipulation training. Through a sampling sphere above the foot contact plane, it \rev{decouples the end-effector goal from base height, roll, and pitch fluctuations}.
    \item To mitigate \rev{catastrophic forgetting of whole-body coordination in multi-terrain training}, we develop a two-stage learning framework with dual-policy distillation, allowing the legged manipulator to possess both whole-body coordination and perceptive traversal capability.
    \item We validate our policy via extensive simulation and real-world experiments. It remarkably expands the reachable workspace, reduces the unexpected collision, and improves end-effector tracking accuracy of cross-terrain loco-manipulation.
\end{itemize}

\section{Related Work}

\begin{figure*}[t]
    \centering
    \includegraphics[width=0.87\linewidth]{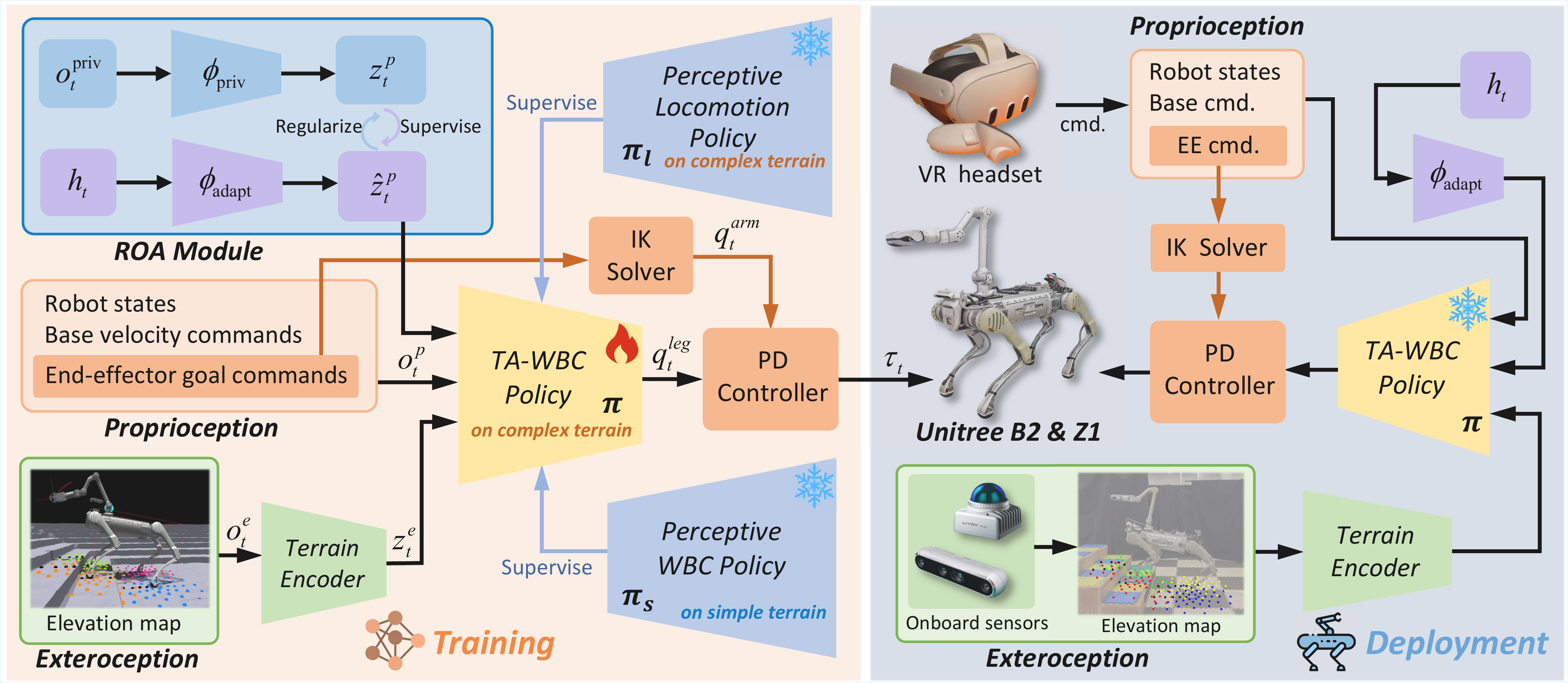}
    \vspace{-10pt}
    \caption{Overview of the TA-WBC framework. TA-WBC is a unified perceptive whole-body controller over complex terrains. First, we train a \rev{perceptive whole-body loco-manipulation expert $\pi_s$ on quasi-level terrain}, and a perceptive \rev{cross-terrain locomotion expert $\pi_l$} on complex terrains with randomized arm joint configurations. After that, \rev{these two experts jointly supervise the unified student policy $\pi$, enabling it} to maintain both the terrain traversability of $\pi_l$ and the whole-body coordination of $\pi_s$ over various terrains. \rev{During deployment, only the unified student policy $\pi$ is executed.} The base and arm commands are sampled uniformly in training stage, while in deployment they are obtained by the VR headset.}
    \label{fig:overview}
    \vspace{-15pt}
\end{figure*}

\subsection{Learning-based Legged Locomotion}
Traditional model-based legged locomotion methods, such as MPC \cite{loco-mpc1, loco-mpc2}, are stable in structured environments but sensitive to unmodeled disturbances. Consequently, deep reinforcement learning (RL) method has become a mainstream alternative.
RL-based locomotion is broadly divided into two directions: blind and perceptive approaches. Blind policies rely on proprioception only, adapting reactively through collision responses and posture variations \cite{wtw, walk-in-min} or implicit terrain imagination \cite{blind1}. Although robust on mildly uneven terrain, they lack anticipation, so the collisions can degrade control precision or trap the robot in hazardous terrain.
Perceptive policies instead use depth images to detect platforms and overhanging obstacles \cite{cheng2024extreme, zhuang2023robot}, or elevation maps to reconstruct continuous 2.5D surroundings \cite{attention, wang2025beamdojo}, enabling robust multidirectional traversal of complex environments.

Although legged locomotion has become increasingly robust, existing works mainly focus on cross-terrain locomotion, overlooking the whole-body loco-manipulation potential over diverse terrains of legged manipulators.

\subsection{Legged Loco-Manipulation}
Coordinated legged loco-manipulation is challenging for high-dimensional joint states and deep coupling between the floating base and the onboard robotic arm. Rather than the decoupled control frameworks, unified architectures are becoming the mainstream method to solve this problem. 
The majority of traditional studies still relied on MPC and hierarchical QP to implement WBC for legged manipulators \cite{spotmini, 2025mpc-wbc,bayes-mpc,Collision-Free-mpc, ALMA}.
While MPC-based methods perform well in specific tasks, they lack sufficient generalization and robustness against external disturbances. To address this, Deep-WBC \cite{dwbc} introduces an RL-based approach by an advantage mixing module. Since then, a series of RL-based WBC frameworks have emerged: 
by integrating visual information, VBC \cite{vbc} and UMI on Legs \cite{umi} enable legged manipulators to execute tasks autonomously; WBC capable of force control are proposed in \cite{force2}; RoboDuet \cite{roboduet} and ReLIC \cite{interlimb} achieves whole-body coordination where the lower body actively assists the robotic arm; and a controller tailored for interactive navigation is developed in \cite{internav}. 
However, most existing whole-body controllers for legged manipulators still lack terrain perception. Although PILOT \cite{pilot} achieves perceptive cross-terrain loco-manipulation for humanoid robots, perceptive WBC remains challenging for quadrupedal legged manipulators.

\section{Methodology}

In loco-manipulation tasks within complex and uneven environments, the primary challenge for legged manipulators lies in mastering high-dimensional control dynamics while simultaneously ensuring the robustness of end-effector tracking.
To address these requirements, we propose TA-WBC, a unified end-to-end reinforcement learning framework tailored for loco-manipulation across various terrains, with the overview shown in Fig.~\ref{fig:overview}. In this section, first, we establish the problem definition, providing a rigorous definition of the state and action spaces. Next, we detail the core components in TA-WBC policy learning framework, highlighting the key modules that enable terrain-aware loco-manipulation.

\vspace{-3pt}

\subsection{Problem Definition}

\subsubsection{State Space Design}
The state space $\mathcal{S}$ is composed of two parts: the proprioceptive observation $o^p_t$ and the perceptive observation $o^e_t$. Specifically, the proprioceptive observation is defined as: 
\begin{equation}
o^p_t = [s^{\text{base}}_t, q_t, \dot{q}_t, a_{t-1}, c_t, s^{\text{foot}}_t,  g_t, \hat{z}_t^p].
\end{equation}
Among them, there are base states $s^{\text{base}}_t \in \mathbb{R}^5$ including roll, pitch, and base angular velocities, joint positions $q_t \in \mathbb{R}^{18}$, joint velocities $\dot{q}_t \in \mathbb{R}^{18}$,  and the last action of the policy $a_{t-1} \in \mathbb{R}^{12}$. Besides, the current commands $c_t=[{v}^{{\text{base}}_d}_t, \omega^{{\text{base}}_d}_t, {p}^{{\text{arm}}_d}_t]$, denoting desired base linear and angular velocities, and \rev{the desired 6D end-effector pose} in the arm frame, respectively. $s^{\text{foot}}_t \in \{0,1\}^4$ denotes the foot contact state, where 1 indicates contact and 0 indicates swing phase. Inspired by \cite{gait,wtw}, we design $g_t \in \mathbb{R}^5$ to provide gait timing reference as a part of observation for a more stable walking pattern. Lastly, $\hat{z}_t^p \in \mathbb{R}^{20}$ represents the environment extrinsic physics, which is predicted by an adaptation encoder from a history of proprioceptive observations.

The exteroceptive observation $o^e_t \in \mathbb{R}^{50\times4}$ corresponds to an elevation map around the legged robot. Following the work of \cite{eth-perceptive-loco}, first, we build a 2.5D robot-centric map from pointcloud perception, then sample an equal amount of points around every foot.

In conclusion, the complete observation of the policy is:
\begin{equation}
o_t=[ o^p_{t-H}, o^p_{t-H+1} ..., o^p_{t-1}, o^p_t, o^e_t],
\end{equation}
where $H = 10$ is the history horizon.

\subsubsection{Action Space Design}
Our controller is capable of controlling the legged robot and the mounted arm simultaneously. \rev{The RL policy outputs only the desired positions of the 12 leg joints, $q^{\text{leg}_d}_t\in\mathbb{R}^{12}$, while the desired arm joint positions $q^{\text{arm}_d}_t\in\mathbb{R}^{6}$ are computed independently by an inverse kinematics (IK) solver from the desired end-effector pose.} Details about arm control can be found in Section \ref{arm control}.

\subsection{Policy Learning}

\begin{figure}[t]
    \centering
    \vspace{-3pt}
    

    \subfloat[Visualization]{
        \includegraphics[width=0.46\linewidth]{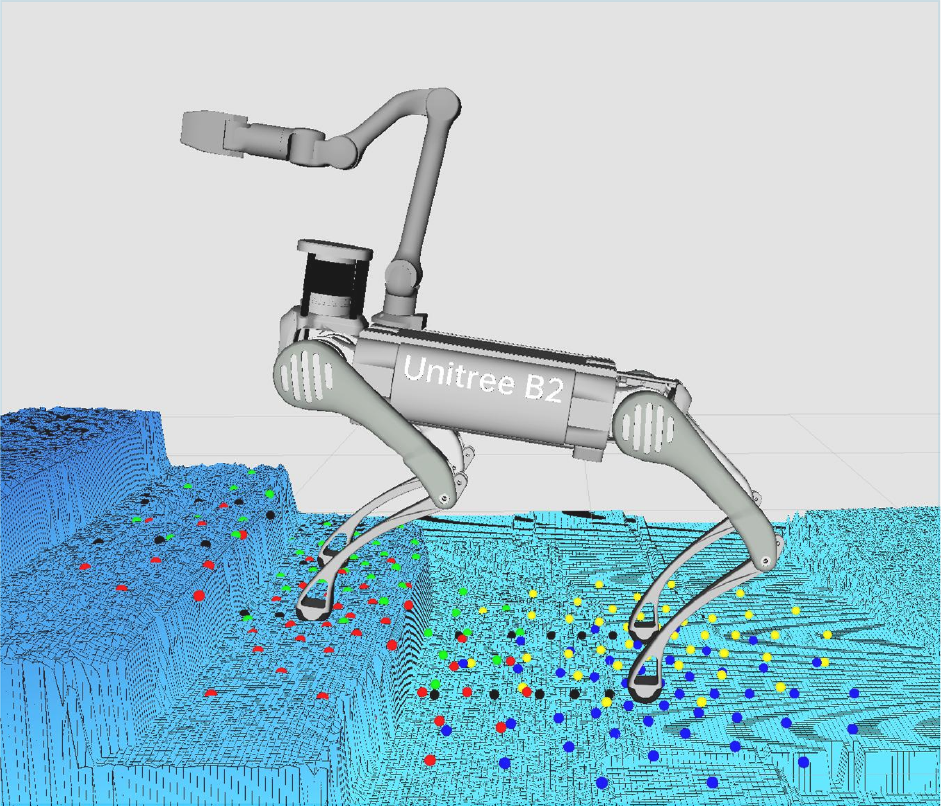}
        \label{fig:200points-b}
    }
    \hfill
    \subfloat[Real-world]{
        \includegraphics[width=0.46\linewidth]{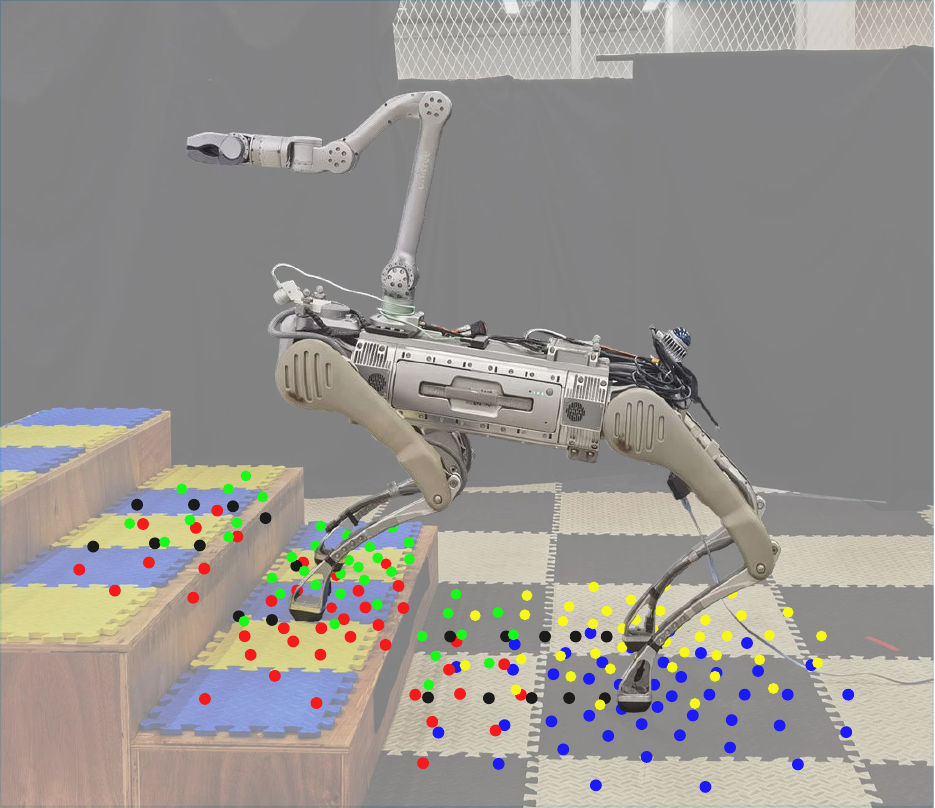}
        \label{fig:200points-c}
    }
    \vspace{-3pt}
    \caption{Details of height sampling points for exteroceptive observation $o^e_t$ in real-world experiment. 
    Samples in different colors correspond to different feet, and 
    samples in direction of $\theta=0$ is represented by black points.}
    \label{fig:200points}
    \vspace{-12pt}
\end{figure}

As shown in Fig. \ref{fig:overview}, 
TA-WBC receives proprioceptive observation $o^p_t$, exteroceptive latent $z^e_t$, and the latent environmental dynamics $\hat{z}_t^p$ inferred from the proprioceptive history $h_t$. During training, two different policies will together supervise the output of TA-WBC, simultaneously enabling robust cross-terrain locomotion and expansive whole-body motion with a unified policy $\pi$. In this section, we will detail the critical modules in policy learning.


\subsubsection{Regularized Online Adaptation}
To achieve partially observable whole-body control \rev{and robust sim-to-real transfer}, we utilize the regularized online adaptation (ROA) \cite{dwbc} framework, consisting of a single actor-critic network and two complementary encoders. The privileged encoder $\phi_{\text{priv}}$ extracts a physics latent vector $z_t^p$ from the privileged information $o_t^{\text{priv}}$, including actual friction, external payload, and motor strength. Simultaneously, an adaptation encoder $\phi_{\text{adapt}}$ is trained to implicitly infer latent environmental dynamics $\hat{z}_t^p$ from the proprioceptive history $h_t = [ o^p_{t-H}, o^p_{t-H+1}, \dots , o^p_{t-1}]$ by imitating $z_t^p$, where $H=10$. Additionally, the physics latent vector $z_t^p$ is regularized to avoid large deviation from $\hat{z}_t^p$. 

\subsubsection{\texorpdfstring{\rev{Hierarchical Exteroceptive Encoder}}{Hierarchical Exteroceptive Encoder}}

Conventional terrain encoder typically relies on base-centric height grip maps, where elevation samples are collected in the base frame. However, this representation suffers from significant spatial misalignment; as the robot maneuvers, the relative transformation between the base-anchored grid and the swing foot remains highly dynamic, often resulting in insufficient perceptive resolution around the feet, which is critical to avoid unexpected collision during traversing terrains. Additionally, this design is always high-dimensional and coupled with an expansive flat MLP, \rev{which can reduce learning efficiency}.

To overcome this limitation, \rev{we choose a multi-ring sampling strategy inspired by \cite{eth-perceptive-loco} to capture dense terrain geometry around each foot, which is critical for preventing lateral foot collisions during cross-terrain loco-manipulation.}
As shown in Fig. \ref{fig:200points}, for each foot $k \in \{1,2,3,4\}$, we sample the terrain heights at $N_k=50$ points arranged in $M=5$ concentric rings centered at the foot's vertical projection. The samples are distributed along multiple concentric circles surrounding each foot, with radii $r = \{0.07, 0.14, 0.21, 0.30, 0.42\}$ m, containing $n =  \{6, 8, 10, 12, 14\}$ points respectively. They are distributed along the angular dimension from $\theta \in [0,2\pi)$, where the direction of $\theta=0$ is aligned with the forward heading of the quadruped base in the horizontal plane. 

This non-uniform density provides higher resolution around each foot while decoupling exteroception from the base frame, enhancing collision-free locomotion over \rev{terrains with height discontinuities, such as stairs and steps}. 
To exploit this structured sampling pattern, a hierarchical encoder first processes each concentric ring independently using circularly padded 1D convolutions. For each foot $k$, the resulting features $M^f_k$ are flattened into $f_k \in \mathbb{R}^{\sum n_i \times n_c}$ and mapped by an MLP to $z^e_k \in \mathbb{R}^{20}$, where $\sum n_i=50$ and $n_c=4$ is the number of convolution output channels.
The encoder is shared across all four feet, yielding the final latent vector $z^e = [z^e_1, z^e_2, z^e_3, z^e_4] \in \mathbb{R}^{80}$. 

This \rev{hierarchical encoder} can enhance training efficiency by explicitly exploiting not only the periodic nature of the circular sampling pattern, but also the inherent morphological symmetry of the quadruped robot to reduce the parameter volume. \rev{Compared with a flat MLP that treats all elevation samples as an unstructured vector, it preserves the circular neighborhood and per-foot local terrain geometry, thereby improving parameter and learning efficiency as well as terrain feature extraction around each foot.}

\subsubsection{End-Effector Goal Sampling}

Most legged loco-manipulation methods assume flat terrain and uniformly sample end-effector targets in a sphere centered at a fixed ground-relative height \cite{vbc}. \rev{Although consistent on flat terrain, this approach ignores terrain-induced base height, roll, and pitch variations. While the method in \cite {eth-ee-tracking} extends end-effector pose sampling on uneven terrain, it explicitly excludes locomotion and requires a static support base configuration. Its sampling sphere is constructed in the fixed base frame, } making it unsuitable for continuous and stable goal sampling during loco-manipulation.

\begin{figure}[t]
    \centering
    \vspace{7pt}
    \includegraphics[width=0.75\linewidth]{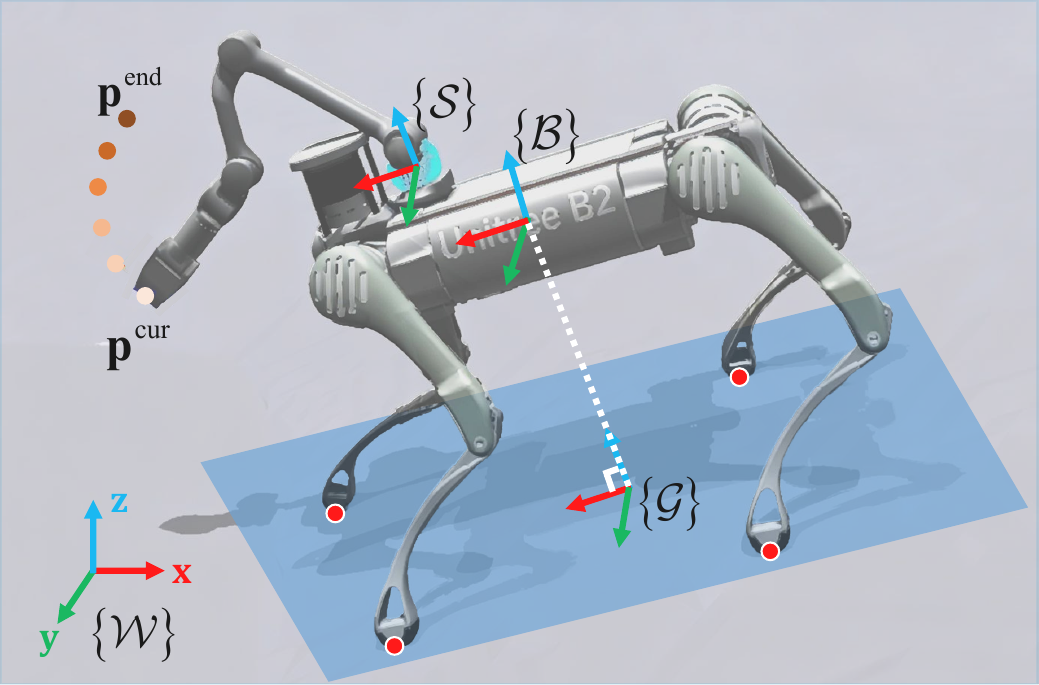}
    \vspace{-4pt}
    \caption{Details of the construction of the FCP and the sampling sphere coordinate. The foot contact points are colored in red, and the FCP is represented by a blue plane.}
    \label{fig:fcp}
    \vspace{-15pt}
\end{figure}

\rev{To enable continuous end-effector command sampling during cross-terrain loco-manipulation, we construct a spherical sampling frame $\mathcal{S}$ at a fixed distance from the foot contact plane (FCP), thereby decoupling the goal from base height, roll, and pitch variations.} As shown in Fig. \ref{fig:fcp}, the FCP is fitted by least squares to the four foot contact points and updated only when all feet have valid ground contact, defined by a vertical contact force markedly greater than its horizontal components. It is represented by a unit normal $\mathbf{n}_f \in \mathbb{R}^3$ and offset $d_f \in \mathbb{R}$, satisfying $\|\mathbf{n}_f\| = 1$ and $\mathbf{n}_f \cdot \mathbf{x} + d_f = 0$.

Let $\mathcal{W}$ and $\mathcal{B}$ denote the world and base frames. Projecting the base position $\mathbf{p}_b$ onto the FCP along $\mathbf{n}_f$ gives $\mathbf{p}_\mathcal{G}$, while the base $x$-axis is projected onto the plane as
\begin{equation}
    \mathbf{v}_{\text{proj}} = \mathbf{x}_\mathcal{B} - (\mathbf{x}_\mathcal{B} \cdot \mathbf{n}_f)\mathbf{n}_f.
\end{equation}
Then we construct the ground projection frame $\mathcal{G}$, which is centered at $\mathbf{p}_\mathcal{G}$, with $\mathbf{R}_\mathcal{G}=[\mathbf{x}_\mathcal{G}, \mathbf{y}_\mathcal{G}, \mathbf{z}_\mathcal{G}] \in SO(3)$ defined by
\begin{equation}
    \mathbf{z}_\mathcal{G} = \mathbf{n}_f, \mathbf{x}_\mathcal{G} = \frac{\mathbf{v}_{\text{proj}}}{\|\mathbf{v}_{\text{proj}}\|},     \mathbf{y}_\mathcal{G} = \mathbf{z}_\mathcal{G} \times \mathbf{x}_\mathcal{G}.
\end{equation}

Finally, the sampling frame $\mathcal{S}$ shares the orientation of $\mathcal{G}$ and is translated by a fixed offset $\mathbf{p}_c$ expressed in $\mathcal{G}$. At every time interval $T_i$, the end-effector goal $\mathbf{p}^{\text{end}}$ is sampled uniformly within a sphere of radius $r_s=0.8$\,m centered at $\mathcal{S}$. Following \cite{vbc}, it is interpolated with the current pose $\mathbf{p}^{\text{cur}}$ to obtain $\mathbf{p}^{\text{cmd}}_t$:
\begin{equation}
    \mathbf{p}_t^{\text{cmd}} = \frac{t}{T_i} \mathbf{p}^{\text{cur}} + \left(1 - \frac{t}{T_i}\right) \mathbf{p}^{\text{end}}, 
    \quad t \in [0, T_i].
\end{equation}

With additional terrain and base-trunk collision checks, FCP-based sampling method provides smooth, adaptive and collision-free 6D end-effector pose sampling across terrains and improves training efficiency.

\subsubsection{Policy Distillation}\label{distill}
To ensure the robot can perform collision-free loco-manipulation across various terrains, while simultaneously actuating whole-body joints to expand end-effector's workspace, we employ a two-stage training method supervised by two \rev{expert} policies.

In the first stage, we train a \rev{perceptive whole-body loco-manipulation expert $\pi_s$ on quasi-level terrain, which shares the same input and output spaces as the student policy $\pi$.} Concurrently,  we train a \rev{perceptive cross-terrain locomotion expert $\pi_l$ on complex terrains with randomized arm joint configurations. It also shares the same input and output spaces as $\pi$, and the only difference is that the end-effector pose command is not randomly sampled from the sphere, but computed by forward kinematics from the sampled arm configuration.}
In the second stage, we relax the mixing ratio introduced in \cite{dwbc}, continuing the training based on $\pi_s$ without any reward scale modification. The unified policy $\pi$ is trained under the supervision of both $\pi_s$ and $\pi_l$ using PPO, augmented with a terrain-conditioned distillation loss defined as:
{\small
\begin{equation}
    \mathcal{L}_{\text{distill}} = \mathbb{E}_{s} \left[ \mathbb{I}(s) D_{\text{KL}}(\pi_s \parallel \pi) + (1 - \mathbb{I}(s)) D_{\text{KL}}(\pi_l \parallel \pi) \right],
\end{equation}
}
where $\mathbb{I}(s)$ is \rev{a training-time indicator} equal to $1$ when on quasi-level planes or received standing command for base, otherwise equal to $0$. $D_{\text{KL}}$ represents the Kullback-Leibler divergence between the continuous action distributions of the respective expert and the unified student policy $\pi$.

\rev{This hard gating assigns supervision according to each expert's competence: $\pi_s$ provides whole-body coordination references, while $\pi_l$ provides collision-free leg-motion references on complex terrains, which is important for base stability and end-effector tracking accuracy. In the second stage, this distillation loss works jointly with the  rewards in Section.~\ref{details} to enable cross-terrain loco-manipulation.}

\subsubsection{Arm Control}\label{arm control} According to \cite{dwbc}, a pure whole-body RL policy struggles in accurately tracking the end-effector orientation. Although a keypoint-based representation of orientation is used in \cite{eth-ee-tracking} and \cite{3dbox} for better tracking, it would increase the dimension of the goal command and introduce additional reward functions.
For convenience, we employ the damped least squares method to solve the inverse kinematics of the robot arm in the world frame. \rev{The FCP-sampled target is transformed from $\mathcal{S}$ into $\mathcal{W}$, while the current end-effector pose is transformed from $\mathcal{B}$ into $\mathcal{W}$. Given the 6D pose error $e$ and the arm Jacobian $J$, the joint increment is computed as $\Delta q = J^\top (J J^\top + \lambda^2 I)^{-1} \mathbf{e}$, which is added to the current arm joint positions to obtain the desired arm joint targets.}

\subsubsection{\texorpdfstring{\rev{Training Details}}{Training Details}}\label{details}\rev{During both training and evaluation, the base linear velocity and yaw-rate commands are uniformly sampled in $[-0.8,0.8]$\,m/s and $[-0.6,0.6]$\,rad/s, respectively. The end-effector position is uniformly sampled in spherical coordinates $(l,p,y)$ aligned with $\mathcal{S}$, where $l\in[0.4,0.8]$\,m, $p\in[-\pi/3,\pi/4]$\,rad, and $y\in[-1.2,1.2]$\,rad; its roll, pitch, and yaw angles are independently sampled from $\mathcal{U}(-\pi/2,\pi/2)$\,rad.} As detailed in Table \ref{tab:rew}, we formulate the reward $r_t$ as a weighted sum of three components: command tracking rewards, base stability rewards, and regularization items. For sim-to-real transfer, we randomize parameters about physical properties (e.g., mass, friction, and actuator strength).

\vspace{-10pt}
\begin{table}[t]
  \renewcommand\arraystretch{1.3}
  \centering
  \small
  \setlength{\tabcolsep}{3.0mm}
  \vspace{8pt}
  \caption{\textsc{Reward Design}}
    \vspace{-5pt}
    \begin{tabular}{cccccc}
    \toprule
    \multicolumn{1}{c}{\textbf{Term}} & \multicolumn{1}{c}{\textbf{Weight}} & \multicolumn{1}{c}{\textbf{Term}} & \multicolumn{1}{c}{\textbf{Weight}} & \multicolumn{1}{c}{\textbf{Term}} & \multicolumn{1}{c}{\textbf{Weight}}\\
    \midrule
    \multirow{6}{*}{}
    $r_{\omega_d}^{\text{base}}$ & $2.0$ & $r_{v_{d}}^{\text{base}}$ & $1.5$ & $r_{p_d}^{\text{arm}}$  & $1.8$ \\
    \midrule
    \multirow{6}{*}{} 
    $r_{\text{height}}^{\text{base}}$ & $-5.0$ & $r_{\text{pitch}}^{\text{base}}$ & $-0.1$ & $r_{{\text{hip}}}^{\text{base}}$ & $-0.8$ \\[5pt]
    $r_{\text{height}}^{\text{feet}}$ & $-4.5$ &
    $r_{{\text{drag}}}^{\text{feet}}$ & $-0.25$ & $r_{{\text{gait}}}^{\text{feet}}$ & $ -3.0 $ \\
    \midrule
    \multirow{6}{*}{} 
    $r_{\tau}$ & $-10^{-7}$ & $r_{{\text{smooth}}}$ & $-0.015$ & $r_{{\text{limit}}}^{\text{dof}}$ & $-1.5$ \\
    \bottomrule
    \end{tabular}%
  \label{tab:rew}%
    \vspace{8pt}

    \centering
    \includegraphics[width=1.0\linewidth]{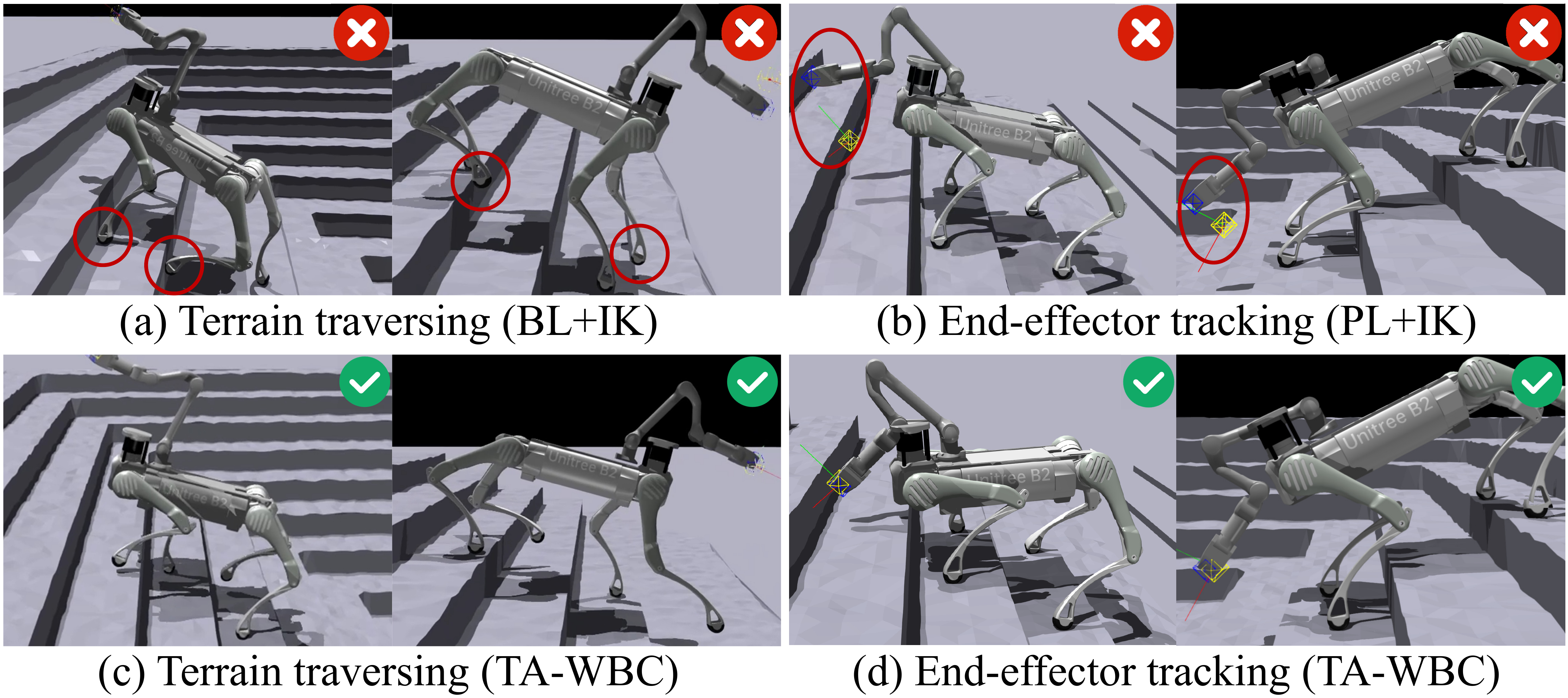}
    \vspace{-14pt}
    \captionof{figure}{\centering Simulation results on various terrains compared with baselines.}
    \label{fig:sim}
    \vspace{-15pt}
\end{table}

\section{Experiments}
\label{sec:experiments}

\subsection{Experiment Setup}



\begin{table*}[t!]
\renewcommand\arraystretch{1.2}
\vspace{7pt}
\caption{\textsc{Simulation results on different terrains.
$\uparrow$: Higher is better; $\downarrow$: Lower is better.}
}
\vspace{-3pt}
\centering
\setlength{\tabcolsep}{2.0mm}
\scriptsize
\begin{tabular}{llcccccccc}
\hline
\textbf{Terrain} & \textbf{Method} 
& \rev{$V_w^s$($\mathrm{m}^3$) $\uparrow$}
& \rev{$V_w^w$($\mathrm{m}^3$) $\uparrow$}
& $l_{\text{ter}}$ $\uparrow$ 
& $E_v$($\mathrm{m/s}$) $\downarrow$  
& $E_{\mathrm{yaw}}$($\mathrm{rad/s}$) $\downarrow$ 
& \rev{$E_{\mathrm{eep}}$($\mathrm{m}$) $\downarrow$}
& \rev{$E_{\mathrm{eeo}}$($\mathrm{rad}$) $\downarrow$}
& $r_c$($\mathrm{\%}$) $\downarrow$ \\
\hline
\multirow{3}{*}{(a) Simple} 
& BL+IK          & 0.710 & \rev{0.678} & - & \textbf{0.032} & \textbf{0.080} & 0.076 & \rev{0.153} & - \\
& Blind WBC      & 0.865 & \rev{0.833} & - & 0.074 & 0.112 & 0.034 & \rev{\textbf{0.086}} & - \\
& TA-WBC (Ours)   & \textbf{0.884} & \rev{\textbf{0.847}} & - & 0.059 & 0.131 & \textbf{0.025} & \rev{0.118} & - \\
\hline
\multirow{3}{*}{(b) Complex} 
& BL+IK          & 0.683 & \rev{0.520} & 6.26 & 0.223 & 0.460 & 0.316 & \rev{0.410} & 18.78 \\
& PL+IK          & 0.683 & \rev{0.578} & 9.00 & \textbf{0.082} & \textbf{0.227} & 0.155 & \rev{0.256} & 0.97 \\
& TA-WBC (Ours)   & \textbf{0.825} & \rev{\textbf{0.714}} & \textbf{9.00} & 0.109 & 0.252 & \textbf{0.042} & \rev{\textbf{0.144}} & \textbf{0.56} \\
\hline
\end{tabular}
\label{tab:baseline}
\vspace{-10pt}
\end{table*}

\begin{figure*}[t!]
    \centering
    
    \vspace{-3pt}
    \subfloat[BL+IK on simple terrain]{
        \includegraphics[width=0.45\linewidth]{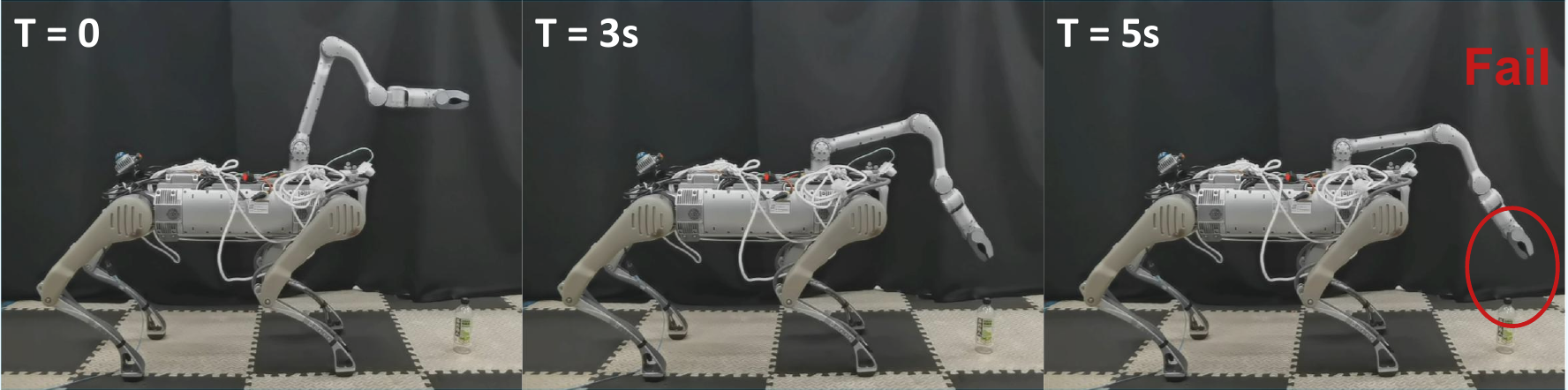}
        \label{fig:stand-ik}
    }
    \subfloat[TA-WBC on simple terrain]{
        \includegraphics[width=0.45\linewidth]{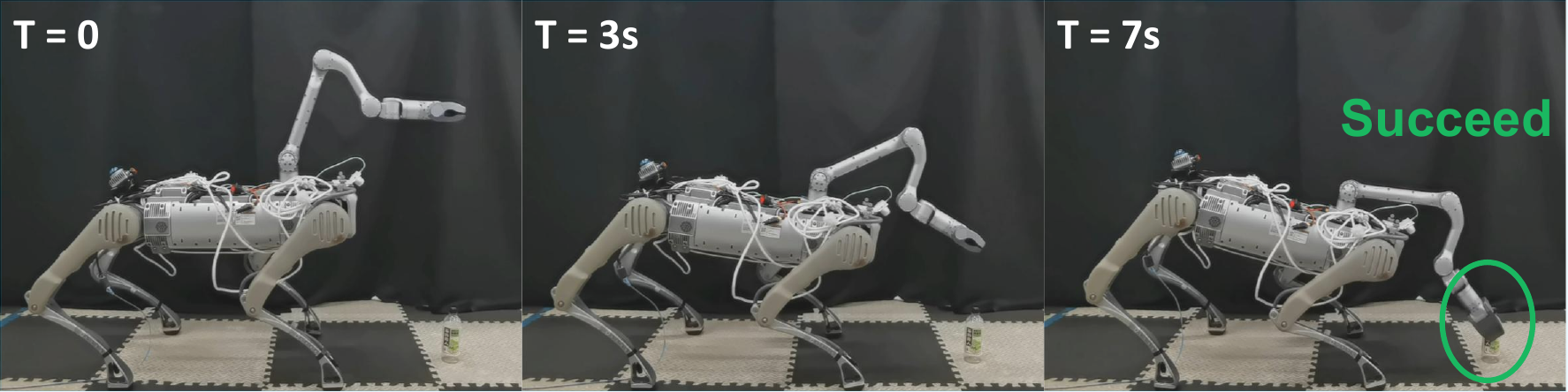}
        \label{fig:stand-wbc}
    } \\ 
    \vspace{-8pt} 

    \subfloat[PL+IK on stairs]{
        \includegraphics[width=0.45\linewidth]{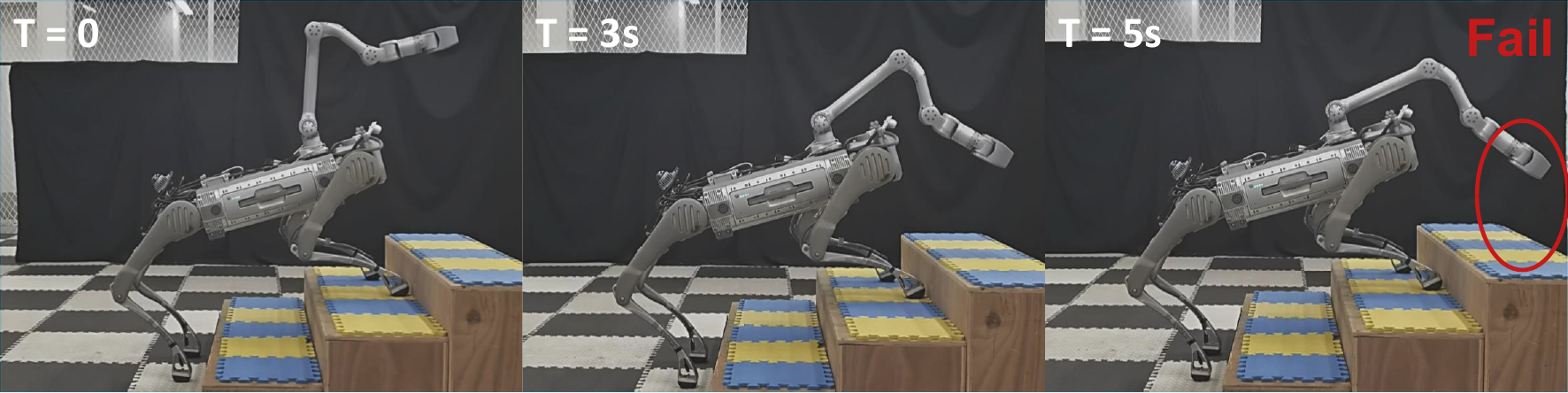}
        \label{fig:stand-ik-stairs}
    }
    \subfloat[TA-WBC on stairs]{
        \includegraphics[width=0.45\linewidth]{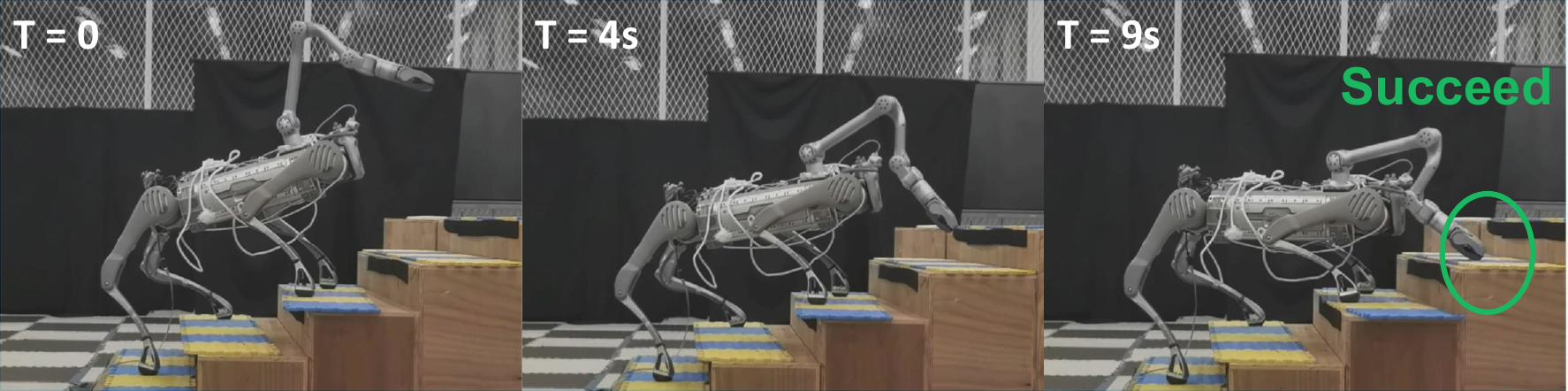}
        \label{fig:stand-wbc-stairs}
    }
    \vspace{-3pt} 

    \caption{\centering Comparison between decoupled controller and TA-WBC when the end-effector goal reaches the ground.}
    \label{fig:stand}
    \vspace{-10pt}
\end{figure*}

\subsubsection{Robot System Setup}
Our hardware platform comprises a Unitree B2 quadruped robot and a Unitree Z1 6-DoF robotic arm equipped with a 1-DoF gripper. To acquire a real-time robot-centric elevation map, a forward-facing Intel RealSense D455 depth camera and a rear-mounted Livox Mid-360 LiDAR are mounted on the B2 torso.

\subsubsection{Evaluation Metrics}
To quantify the performance of the proposed method and baselines, we define the following metrics. 
(1) linear velocity tracking error ($E_v$): the difference between command and actual base linear velocity; 
(2) yaw rate tracking error ($E_{\mathrm{yaw}}$): the difference between command and actual base yaw rate; 
(3) end-effector position tracking error \rev{($E_{\mathrm{eep}}$)}: the Euclidean distance between the target and actual end-effector positions;
\rev{(4) end-effector orientation tracking error ($E_{\mathrm{eeo}}$): the geodesic rotation angle between the target and actual end-effector orientations;}
\rev{(5) standing and walking workspace ($V_w^s$ and $V_w^w$):} subtracted by the volume of the convex hull formed by 1000 reachable end-effector target positions. Specifically, a target position is considered reachable if the end-effector successfully reaches it with \rev{$E_{\mathrm{eep}} \le 0.03$\,m} and \rev{$E_{\mathrm{eeo}} \le 0.2$\,rad} without collision; 
(6) unexpected contact rate ($r_c$): the frequency of horizontal impacts or accidental slips occurring on the four feet during traversing complex terrains, especially on stairs; 
(7) maximum available terrain level ($l_{\text{ter}}$): the highest terrain level traversable by the robot. Traversability is defined as achieving a success rate exceeding 95\% with $E_v \le 0.1 $\,m/s. $l_{\text{ter}} \in [0,9]$, where 9 represents the maximum stair height, discrete step height, and slope inclination angles. 
\rev{Except for ($V_w^s$ and $V_w^w$), all metrics are measured during simultaneous loco-manipulation. In each evaluation, we randomly sample 10 distinct 6D end-effector target poses away from the default arm configuration. The base and end-effector commands follow the same sampling distributions as in training (Section~\ref{details}) and are activated simultaneously.}

\begin{table*}[t]
\centering
\scriptsize
\vspace{4pt}
\renewcommand\arraystretch{1.2}
\caption{\textsc{Ablation results of key modules over complex terrains.}}
\vspace{-4pt}
\label{tab:ablation}
\resizebox{\textwidth}{!}{%
\begin{tabular}{ccccccccccccccc} 
\toprule
\textbf{Method} & \textbf{Exter.} & \textbf{Foot.} & \rev{\textbf{Hie.}} & \rev{\textbf{FCP}} & \textbf{Distill.} & \rev{$V_w^s$($\mathrm{m}^3$) $\uparrow$} & \rev{$V_w^w$($\mathrm{m}^3$) $\uparrow$} & $l_{\text{ter}}$ $\uparrow$ & $E_v$($\mathrm{m/s}$) $\downarrow$ & $E_{\mathrm{yaw}}$($\mathrm{rad/s}$) $\downarrow$ & \rev{$E_{\mathrm{eep}}$($\mathrm{m}$) $\downarrow$} & \rev{$E_{\mathrm{eeo}}$($\mathrm{rad}$) $\downarrow$} & \rev{$\Delta\theta_{\mathrm{pitch}}$($\mathrm{rad}$) $\uparrow$} & $r_c$($\mathrm{\%}$) $\downarrow$ \\
\midrule
(a) & \textcolor[HTML]{B22222}{\ding{55}} & \textcolor[HTML]{B22222}{\ding{55}} & \textcolor[HTML]{B22222}{\ding{55}} & \textcolor[HTML]{228B22}{\ding{51}} & \textcolor[HTML]{B22222}{\ding{55}} & 0.766 & \rev{0.545} & 5.57 & 0.205 & 0.431 & 0.289 & \rev{0.372} & \rev{0.154} & 24.37 \\
(b) & \textcolor[HTML]{228B22}{\ding{51}} & \textcolor[HTML]{B22222}{\ding{55}} & \textcolor[HTML]{228B22}{\ding{51}} & \textcolor[HTML]{228B22}{\ding{51}} & \textcolor[HTML]{228B22}{\ding{51}} & 0.814 & \rev{0.677} & 8.76 & 0.127 & 0.309 & 0.070 & \rev{0.149} & \rev{0.478} & 2.57 \\
(c) & \textcolor[HTML]{228B22}{\ding{51}} & \textcolor[HTML]{228B22}{\ding{51}} & \textcolor[HTML]{B22222}{\ding{55}} & \textcolor[HTML]{228B22}{\ding{51}} & \textcolor[HTML]{228B22}{\ding{51}} & 0.809 & \rev{0.693} & 8.20 & 0.126 & 0.274 & 0.103 & \rev{0.177} & \rev{0.506} & 4.20 \\
\rev{(d)} & \textcolor[HTML]{228B22}{\ding{51}} & \textcolor[HTML]{228B22}{\ding{51}} & \textcolor[HTML]{228B22}{\ding{51}} & \textcolor[HTML]{B22222}{\ding{55}} & \textcolor[HTML]{228B22}{\ding{51}} & \rev{0.763} & \rev{0.589} & \rev{8.57} & \rev{0.182} & \rev{0.358} & \rev{0.135} & \rev{0.310} & \rev{0.547} & \rev{0.65} \\
(e) & \textcolor[HTML]{228B22}{\ding{51}} & \textcolor[HTML]{228B22}{\ding{51}} & \textcolor[HTML]{228B22}{\ding{51}} & \textcolor[HTML]{228B22}{\ding{51}} & \textcolor[HTML]{B22222}{\ding{55}} & 0.742 & \rev{0.606} & 8.73 & 0.118 & 0.365 & 0.161 & \rev{0.274} & \rev{0.122} & 0.88 \\
TA-WBC & \textcolor[HTML]{228B22}{\ding{51}} & \textcolor[HTML]{228B22}{\ding{51}} & \textcolor[HTML]{228B22}{\ding{51}} & \textcolor[HTML]{228B22}{\ding{51}} & \textcolor[HTML]{228B22}{\ding{51}} & \textbf{0.825} & \rev{\textbf{0.714}} & \textbf{9.00} & \textbf{0.109} & \textbf{0.252} & \textbf{0.042} & \rev{\textbf{0.144}} & \rev{\textbf{0.611}} & \textbf{0.56} \\
\bottomrule
\end{tabular}
}

\begin{flushleft}
    {\textbf{Remark:} (a) Removing all exteroceptive inputs; (b) replacing foot-centric elevation sampling with a base-centric grid; (c) replacing the \rev{hierarchical} terrain encoder with a flat MLP; \rev{(d) replacing FCP-based end-effector goal sampling with base-frame sampling;} (e) removing policy distillation.}
\end{flushleft}
\vspace{-17pt}
\end{table*}

\subsection{Simulation Results}

\subsubsection{Comparison with Baselines}
We compare our method against baselines in both simple plane and complex non-flat scenarios with slopes, steps, and stairs.

\underline{\textit{\textbf{Simple Terrain:}}}
We first verify the performance on simple quasi-horizontal terrain, which is approximately a level plane without significant irregularities.We compare TA-WBC against the following algorithms:
(1) \textbf{Blind Locomotion and IK (BL+IK)}: a proprioception-based locomotion policy with a decoupled IK solver for the robotic arm;
(2) \textbf{Blind Whole-Body Control (Blind WBC)}: a whole-body controller for legged manipulators with proprioception only, which is trained based on \cite{vbc}.

As shown in Table \ref{tab:baseline}(a), despite the remarkably increased input dimensionality from the additional terrain perception, TA-WBC maintains excellent performance similar to Blind WBC on flat ground.
\rev{Compared with the decoupled BL+IK baseline, $E_v$ and $E_{\text{yaw}}$ of TA-WBC slightly increases, because the unified policy must balance base-command tracking with simultaneous end-effector tracking, leading to a trade-off for whole-body coordination. Nevertheless, TA-WBC substantially improves the reachable workspace and end-effector tracking accuracy, which are more critical for loco-manipulation.}

\underline{\textit{\textbf{Complex Terrain:}}}
To verify the terrain adaptability and loco-manipulation capability of TA-WBC, we compare it against the following methods on complex terrains including quasi-plane, ascending and descending slopes, staircases, and \rev{steps}: 
(1) \textbf{Blind Locomotion and IK (BL+IK)}: a proprioception-based cross-terrain locomotion policy with a decoupled IK solver for the arm;
(2) \textbf{Perceptive Locomotion and IK (PL+IK)}: a perceptive cross-terrain locomotion policy with a decoupled IK solver for the arm.

As shown in Table \ref{tab:baseline}(b) and Fig. \ref{fig:sim}(a), the robot with BL+IK fails to lift its feet proactively before horizontal contact occurs, resulting in the highest $r_c$ and the largest \rev{$E_{\mathrm{eep}}$}. The robot with PL+IK is able to anticipate terrain and avoid stumbles, however, as shown in Fig. \ref{fig:sim}(b), maintains a relatively fixed base pitch and leads to a limited reachable workspace. \rev{In contrast, TA-WBC avoids unexpected foot contacts (Fig.~\ref{fig:sim}(c)) and coordinates the leg and base motions to expand the reachable workspace while the robot is standing or walking (Fig.~\ref{fig:sim}(d)). Over complex terrains, it is obvious that $V_w^w$ is much smaller than $V_w^s$. The reason is that reaching extreme low  end-effector targets requires deeper flexion of the front legs and a forward body posture with a large base pitch, which tends to shift the CoM and contact load toward the front legs, significantly increasing the required joint torques. During standing, all four feet can maintain contact and support this posture under quasi-static balance. However, in simultaneous loco-manipulation, the controller must repeatedly transfer support, preserve swing-foot clearance, and reject terrain-induced disturbances. It therefore prioritizes successful collision-free terrain traversal by limiting extreme whole-body postures, resulting in a smaller reachable workspace.}
\subsubsection{Ablation Study}
\label{ssub:ablation}
We perform ablation studies to validate the contribution of each module, as detailed in Table \ref{tab:ablation}. \rev{We introduce the base pitch range $\Delta\theta_{\mathrm{pitch}}$, to quantify how extensively the controller coordinates the base to assist manipulation.}

With a pure proprioceptive WBC over complex terrains (a), the robot is unable to perceive upcoming terrain and react in advance, resulting in the worst performance across all metrics, which validate the indispensable nature of exteroception. By replacing foot-centric multi-ring sampling with a base-centric grip one (b), or employing a flat MLP rather than our \rev{hierarchical} architecture (c), $l_{\text{ter}}$ decreases and foot stumble $r_c$ increases by at least 400\%. \rev{The replacement of FCP-based end-effector goal sampling with base-frame sampling (d) leads to a significant increase in $E_{\mathrm{eep}}$ and $E_{\mathrm{eeo}}$, which verify the advantage of our method compared with the one in \cite{eth-ee-tracking}.} Moreover, the removal of policy distillation module (e) leads to \rev{training conflict between cross-terrain locomotion and whole-body motion, resulting in an almost fixed pitch angle of robot base with the least $\Delta\theta_{\mathrm{pitch}}$, therefore significantly reducing $V_w^s$, $V_w^w$ and leading to a limited reachable workspace.} These results validate the contributions of individual components in enhancing cross-terrain stability and expanding workspace.
\subsection{Real-World Deployment}
\label{subsec:real}
We directly deploy the trained policy in a real-world robot system via zero-shot transfer. The whole-body policy $\pi$ outputs the target joint positions for 12 leg joints at 50\,Hz, which are subsequently transmitted to PD controllers for motors that run at 500\,Hz. For the acquisition of real-time terrain perception, we utilize the onboard depth camera and LiDAR to construct the elevation map with the framework in \cite{ele-cupy}. For convenience, the high-level base and end-effector commands are obtained by VR teleoperation from a Meta Quest 3.

\subsubsection{Standing Whole-Body Motion}

As shown in Fig. \ref{fig:stand}(a), when the end-effector target pose is very low (even when reaching the ground), the legged manipulator with the decoupled policy fails to reach due to the original limitation of the arm workspace. However, with TA-WBC shown in Fig. \ref{fig:stand}(b), the robot is able to adjust leg joints and grasp the bottle on flat ground with a height less than 10\,cm. Such coordination enables substantial base pitch adjustments to assist the robotic arm, thereby remarkably expanding the arm workspace. Furthermore, in the stair scenario, we manually stop the robot midway while ascending. As shown in Fig. \ref{fig:stand}(d), its hind left and right feet even stand on two different steps, leading to a non-zero roll and pitch base pose. Although it is challenging to keep balance in that case, our method still enables the end-effector to directly reach the ground, with a dynamic variation on pitch exceeding 25 degrees.

\subsubsection{Long-Horizon Loco-Manipulation}
\begin{figure}[t]
    \centering
    \includegraphics[width=\linewidth]{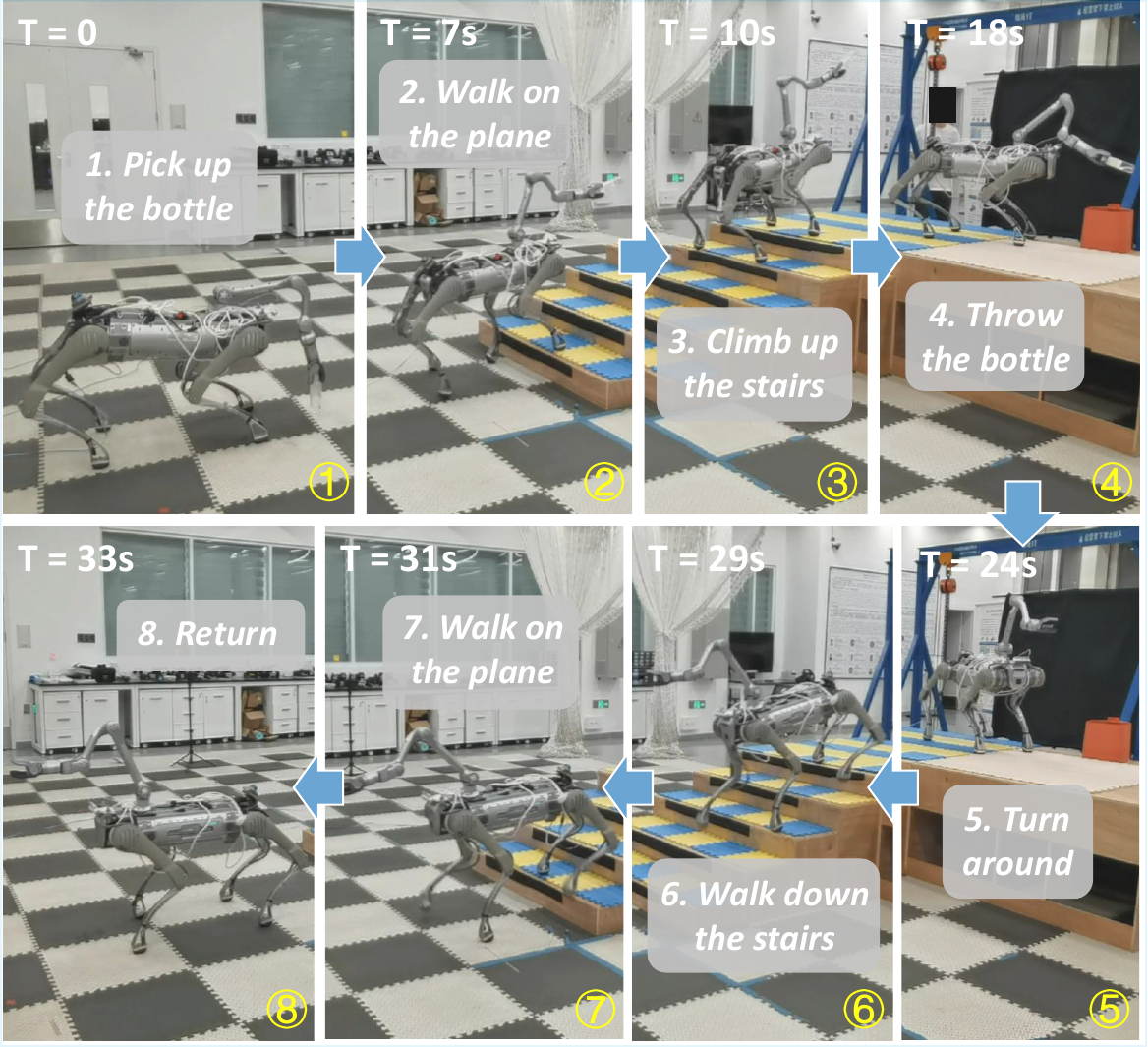}
    \vspace{-15pt}
    \caption{Overview of the continuous long-horizon cross-terrain loco-manipulation task.}
    \label{fig:long}
    \vspace{-14pt}
\end{figure}

To verify the terrain adaptability and loco-manipulation capability of TA-WBC in the real World, we design a long-horizon cross-terrain loco-manipulation task shown in Fig. \ref{fig:long}. Specifically, the robot is teleoperated to execute a series of tasks: first, start from flat plane and pick up the bottle on the ground \rev{with whole-body joint coordination}; then, climb up the stairs consisting of five steps with a height of 17\,cm \rev{, while the onboard arm is moving from left to right}; after that, approach the trash bin placed on the platform and throw the bottle in; finally, turn around on the platform and walk downstairs, returning to the starting zone. 

In this real-world case, our method enables the robot to walk on stairs at a speed of up to 0.8\,m/s without any unexpected collision between feet and terrain, demonstrating strong flexibility and high security. We repeatedly restart and test by setting different starting zones, diverse grasping targets, and various velocity commands, the results show that the legged manipulator with TA-WBC can successfully finish this long-horizon loco-manipulation task 5 times in a row without any falls or stumbles, verifying its terrain adaptability and end-effector tracking capability.

\section{Conclusion}
In this paper, we propose TA-WBC, a terrain-aware unified end-to-end RL framework for legged loco-manipulation over various terrains. With \rev{a hierarchical exteroceptive encoder for foot-centric terrain perception, an FCP-based method for end-effector command sampling, and assistance of policy distillation}, TA-WBC first achieves expansive and cross-terrain whole-body control with one single policy. Simulation and real-world deployment validate its robustness on terrain traversability and expanding end-effector tracking \rev{during cross-terrain loco-manipulation}. Future works include the development of sophisticated high-level policies that seamlessly integrate with TA-WBC as the robust low-level controller, which is expected to enable the legged manipulators to achieve fully autonomous loco-manipulation task execution over various terrains.
\bibliographystyle{IEEEtran}
\bibliography{refs}

\end{document}